%% file: main.tex
\documentclass[sigconf,nonacm,10pt,screen]{acmart}

\AtBeginDocument{%
  }

\usepackage{array}
\usepackage{textcomp}
\usepackage{stfloats}
\usepackage{url}
\usepackage{verbatim}
\usepackage{graphicx}
\usepackage{hyperref}
\usepackage{algpseudocode}
\usepackage{booktabs}
\usepackage{multirow}
\usepackage{threeparttable}
\usepackage{colortbl}
\usepackage{makecell}
\usepackage{tabularx}
\usepackage{wasysym} 
\usepackage{xcolor}       
\usepackage{pifont}       
\usepackage{ragged2e}
\usepackage{enumitem}
\usepackage{appendix}
\usepackage{soul}
\usepackage{multicol}

\setlist[itemize]{noitemsep, topsep=0pt}
\setlist[enumerate]{noitemsep, topsep=0pt}
\setlist[itemize]{leftmargin=0.4cm} 
\setlist[enumerate]{leftmargin=0.4cm} 
\setcopyright{none} 
\usepackage{caption}
\makeatletter
\renewcommand\section{\@startsection{section}{1}{0pt}%
  {1.5ex plus 0.2ex minus 0.1ex}
  {0.8ex plus 0.1ex}
  {\normalfont\Large\bfseries}}

\renewcommand\subsection{\@startsection{subsection}{2}{0pt}%
  {1.2ex plus 0.2ex minus 0.1ex}
  {0.6ex plus 0.1ex}
  {\normalfont\large\bfseries}}

\renewcommand\subsubsection{\@startsection{subsubsection}{3}{0pt}%
  {1ex plus 0.1ex minus 0.1ex}
  {0.5ex plus 0.1ex}
  {\normalfont\normalsize\bfseries}}
\pagebreak[1]
\begin{document}
\title{The Starlink Robot: A Platform and Dataset for Mobile Satellite Communication}
\author{Boyi Liu}
\affiliation{%
  \institution{University College London}
  \country{}
}
\email{bliubd@connect.ust.hk}

\author{Qianyi Zhang}
\affiliation{%
  \institution{University College London}
  \country{} 
}
\email{qianyi.zhang@ucl.ac.uk}

\author{Qiang Yang}
\affiliation{%
  \institution{University of Cambridge}
  \country{}
}
\email{qy258@cam.ac.uk}

\author{Jianhao Jiao}
\affiliation{%
  \institution{University College London}
  \country{}
}
\email{jiaojh1994@gmail.com}

\author{Jagmohan Chauhan}
\affiliation{%
  \institution{University College London}
  \country{}
}
\email{jagmohan.chauhan@ucl.ac.uk
}

\author{Dimitrios Kanoulas}
\affiliation{%
  \institution{University College London}
  \country{}
}
\email{d.kanoulas@ucl.ac.uk}

\input{Sections/0_Abstract}
\maketitle
\textcolor{red}{\textbf{This work was supported by the UKRI Future Leaders Fellowship [MR/V025333/1] (RoboHike). For the purpose of Open Access, the author
has applied a CC BY public copyright license to any Author Accepted
Manuscript version arising from this submission.}}
\input{Sections/1_Introduction}
\input{Sections/2_Related_Works}
\input{Sections/3_System}
\input{Sections/4_Dataset}
\input{Sections/5_Preliminary_Analysis}
\input{Sections/6_Final}

\bibliographystyle{ACM-Reference-Format}
\bibliography{ref}
\end{document}

%% file: Sections/0_Abstract.tex
\begin{abstract}
The integration of satellite communication into mobile devices represents a paradigm shift in connectivity, yet the performance characteristics under motion and environmental occlusion remain poorly understood. We present the Starlink Robot, the first mobile robotic platform equipped with Starlink satellite internet, comprehensive sensor suite including upward-facing camera, LiDAR, and IMU, designed to systematically study satellite communication performance during movement. Our multi-modal dataset captures synchronized communication metrics, motion dynamics, sky visibility, and 3D environmental context across diverse scenarios including steady-state motion, variable speeds, and different occlusion conditions. This platform and dataset enable researchers to develop motion-aware communication protocols, predict connectivity disruptions, and optimize satellite communication for emerging mobile applications from smartphones to autonomous vehicles. In this work, we use \href{https://github.com/clarkzjw/LEOViz/}{LEOViz} for real-time satellite tracking and data collection. The starlink robot project is available at \url{https://github.com/StarlinkRobot}.
\end{abstract}

\keywords{Satellite Communication, Mobile Systems, Robot, Starlink}

%% file: Sections/1_Introduction.tex
\begin{figure}[!htbp]
    \centering
    \includegraphics[width=0.48\textwidth]{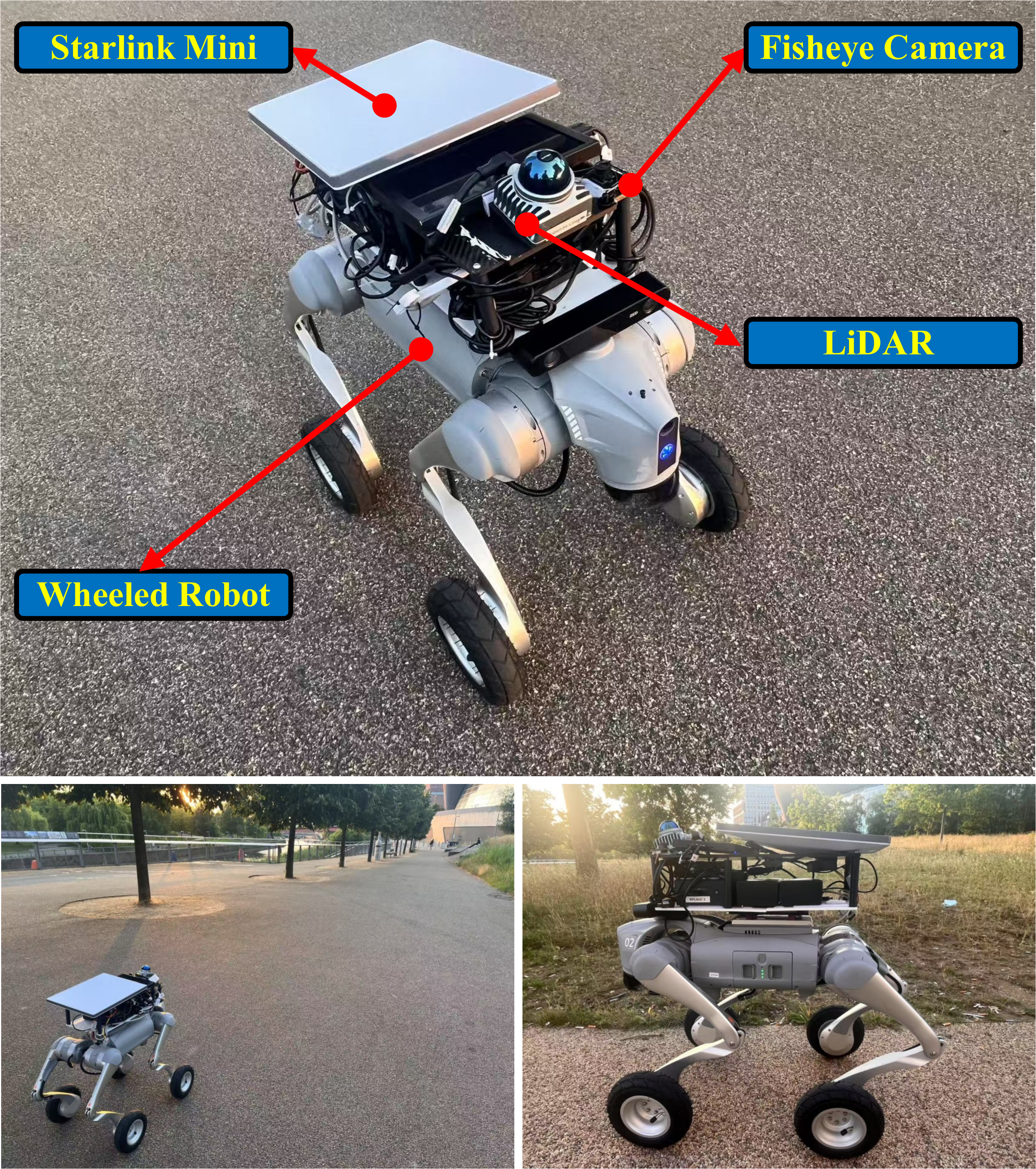}
    \caption{The Starlink Robot platform integrating a Unitree GO2 wheeled robot with Starlink Mini terminal, upward-facing fisheye camera, and Livox Mid-360 LiDAR for comprehensive mobile satellite communication research.}
    \label{fig:firstFigure}
\end{figure}
\section{Introduction}
\label{introduction}
The landscape of global connectivity is undergoing a fundamental transformation. SpaceX's Starlink constellation has deployed over 5,000 satellites, delivering high-speed internet to previously unreachable locations. This success has catalyzed a broader revolution: major technology companies including Apple, Samsung, and Google are racing to integrate satellite communication capabilities directly into consumer smartphones. The promise is compelling – seamless connectivity anywhere on Earth, free from the constraints of terrestrial infrastructure.

Yet this promise faces a critical challenge. Current satellite internet deployments predominantly serve stationary users – homes, businesses, and fixed installations. The Starlink Mini's recent introduction has made portable satellite internet more accessible, but fundamental questions remain unanswered. How does motion affect satellite link quality? What happens when a device moves through environments with varying sky visibility? These questions become urgent as we envision a world where every mobile device, from smartphones to delivery robots to autonomous vehicles, maintains constant satellite connectivity.

The challenge extends beyond simple mobility. Satellite communication operates under fundamentally different constraints than terrestrial networks. A moving device must maintain connection with satellites traveling at 7.5 km/s while simultaneously dealing with local motion and environmental occlusions. Trees, buildings, and even the device's own orientation can disrupt the delicate link between Earth and space. Understanding these dynamics requires more than theoretical models – it demands real-world data collected under controlled yet realistic conditions.

To address this gap, we developed the Starlink Robot shown in Figure~\ref{fig:firstFigure}, a purpose-built platform that brings together mobile robotics and satellite communication. Our approach transforms a Unitree GO2 wheeled robot into a mobile laboratory, equipped with Starlink Mini for connectivity and a suite of sensors to capture the complete context of communication performance. The upward-facing fisheye camera observes sky visibility, the Livox Mid-360 LiDAR maps the surrounding environment, and integrated IMUs track every movement. This comprehensive sensing enables us to correlate communication metrics with physical conditions, revealing how motion and occlusion influence satellite connectivity.

Our contribution extends beyond the platform itself. We present a \textbf{multi-modal dataset} that synchronizes Starlink performance metrics – including latency, upload and download throughput, and signal quality indicators – with high-frequency motion data and environmental observations. Satellite tracking data is collected using LEOViz~\cite{ahangarpour2024trajectory, LEOViz2025}, which provides real-time visualization of satellite positions and connection status. This dataset captures diverse scenarios from steady locomotion to variable speeds, from open sky to heavily occluded urban environments. By releasing both our platform design and collected data, we provide the research community with tools to develop the next generation of mobile satellite communication systems. 

%% file: Sections/2_Related_Works.tex
\section{Related Works}
The rapid deployment of LEO satellite constellations has sparked significant research interest in characterizing their performance. The Starlink academic community, particularly through the University of Victoria's PanLab, has produced comprehensive studies of Starlink's several static performance characteristics~\cite{zhao2024lens, yang2024mobility, zheng2024adaptive, tian2024ebpf, kamel2024starquic, ahangarpour2024trajectory, li2025ftrl, wang2024large, liu2025modeling, zheng2025qter, hu2020directed, yang2021mm, deng2022distance, wang2023high, pan2023measuring, pan2024measuring, zhao2024low}. Among these, LEOViz~\cite{ahangarpour2024trajectory, LEOViz2025} provides real-time visualization and tracking of Starlink satellites, displaying satellite positions, elevation angles, and connection status. In our work, we utilize LEOViz as our data collection and visualization tool, running it alongside our robotic platform to record satellite tracking information during mobile experiments. The satellite data captured by LEOViz enables us to analyze satellite communication performance during movement. Kassem et al. analyzed throughput variations across geographic locations, revealing how latitude affects connection quality due to satellite density differences~\cite{kassem2022browser}. Muhammad et al. investigated weather impacts, demonstrating that rain fade affects Starlink less severely than traditional geostationary satellites due to shorter signal paths~\cite{ullah2025impact}. These foundational studies establish baseline performance metrics but explicitly acknowledge the limitation of stationary measurements.

Recent work has begun exploring mobility scenarios, though primarily in constrained settings. Laniewski et al. conducted preliminary tests with Starlink terminals in vehicles, reporting increased latency variance during highway driving~\cite{laniewski2025measuring}. However, their study lacked synchronized motion data and environmental context, making it difficult to isolate causative factors. SpaceX Maritime~\cite{starlink_maritime} deployments documented by SpaceX show promising performance on ships, but the relatively stable motion and unobstructed ocean views present a best-case scenario that doesn't translate to terrestrial mobile applications. The fundamental challenge remains: no existing work provides the fine-grained, multi-modal data necessary to understand how specific motion patterns and environmental conditions affect satellite communication.

The robotics community has long recognized the value of mobile platforms for wireless communication research. The CRAWDAD repository~\cite{ieee_crawdad} contains numerous datasets from robot-mounted WiFi experiments, demonstrating how controlled mobility can reveal network behavior patterns invisible in static deployments. More recently, researchers have employed drones to map 5G coverage, taking advantage of three-dimensional mobility to characterize cellular networks~\cite{platzgummer2019uav}. Yet these efforts remain confined to terrestrial communication systems~\cite{ITU_M2150_2021}. The unique challenges of satellite communication – including the need for precise sky visibility, the impact of antenna orientation, and the effects of Doppler shift from dual mobility – require purpose-built platforms and measurement methodologies. Our work bridges this gap by adapting mobile robotics techniques specifically for satellite communication research, creating a reproducible platform that others can build upon.

Our work addresses this gap by creating the first dedicated platform for mobile satellite communication research, providing the tools and data necessary to understand this emerging communication paradigm.

%% file: Sections/3_System.tex
\section{System Design}
\begin{figure}[t]
    \centering
    \includegraphics[width=0.48\textwidth]{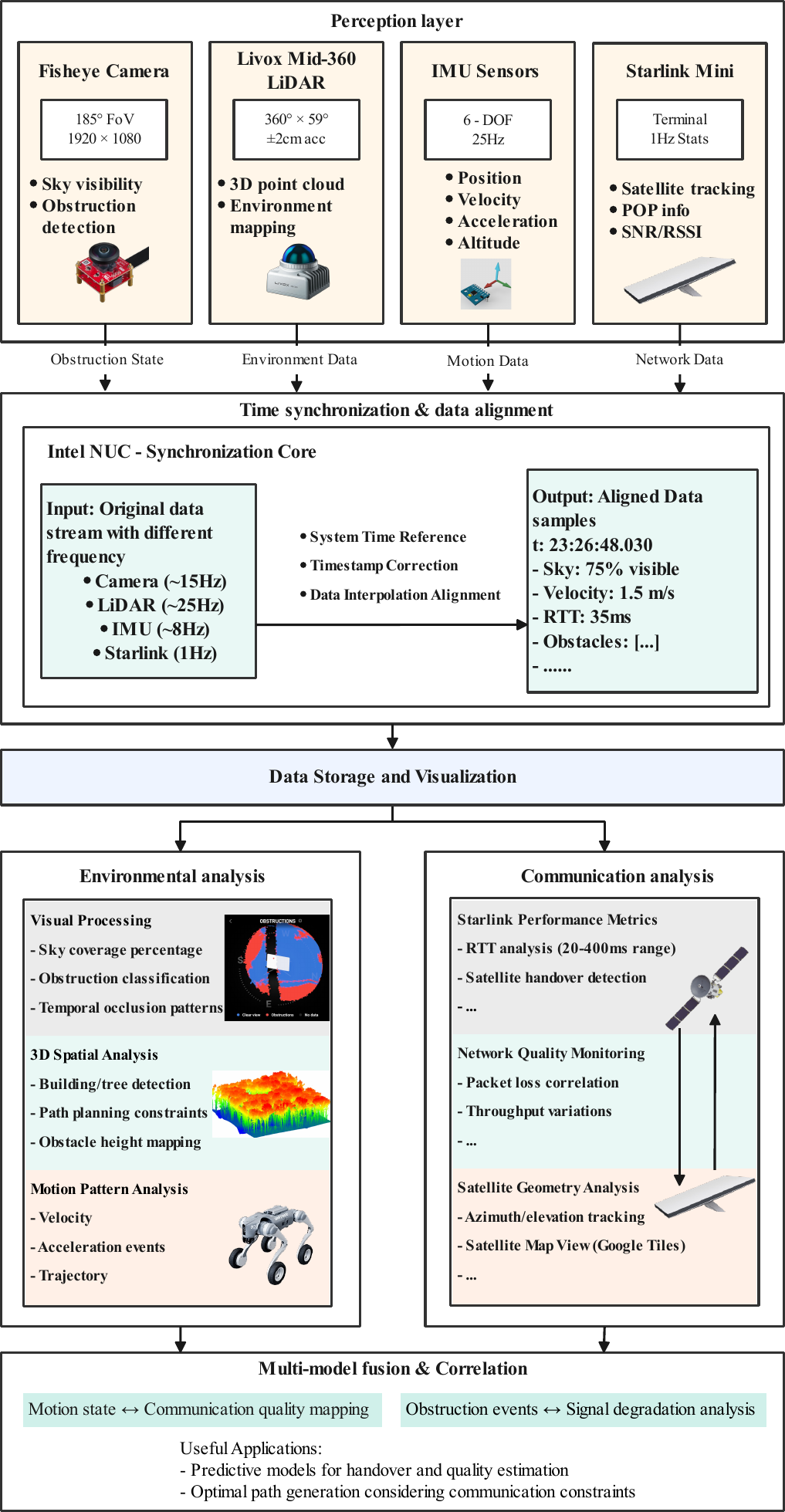}
    \caption{System architecture of the Starlink Robot platform showing multi-modal sensor integration, data synchronization pipeline, and analysis framework for correlating communication performance with environmental and motion context.}
    \label{fig:SystemFramewok}
\end{figure}
Creating a mobile platform for satellite communication research requires careful integration of robotics, networking, and sensing technologies. Our design philosophy prioritizes stability, reproducibility, and comprehensive data collection while maintaining the mobility necessary to explore diverse scenarios. Figure~\ref{fig:SystemFramewok} illustrates the complete system architecture, from sensor integration to data analysis pipelines.

\subsection{Hardware Architecture}
The hardware is shown in Figure~\ref{fig:firstFigure}. The foundation of our system is the Unitree GO2 wheeled robot, chosen for its unique combination of stability and agility. Unlike legged robots that introduce gait-induced vibrations, the wheeled configuration provides smooth motion essential for maintaining satellite links. The GO2's 15kg payload capacity accommodates our full sensor suite while maintaining dynamic performance, and its differential drive system enables precise velocity control from 0.1 to 2.0 m/s – spanning the range from slow walking to jogging speeds.

The Starlink Mini terminal mounts atop the robot via a custom aluminum frame designed to maintain the manufacturer's recommended orientation while minimizing vibration transmission. Power delivery posed an interesting challenge: the Starlink Mini's 12V requirement differs from the robot's 24V battery system. We resolved this with a high-efficiency DC-DC converter that provides stable power while minimizing electromagnetic interference with the sensitive satellite receiver. The compact Intel NUC onboard computer serves as the central data collection hub, running Ubuntu 18.04 with ROS Noetic for sensor coordination and data logging.

Our sensing configuration captures the complete context needed to understand communication performance. The Livox Mid-360 LiDAR provides 360-degree environmental mapping with 0.05m ranging accuracy, essential for identifying potential occlusion sources before they impact connectivity. Mounted directly above the Starlink terminal, our upward-facing fisheye camera captures a 185-degree field of view of the sky. This placement allows us to correlate visible sky percentage with signal quality in real-time. The robot's integrated IMU system provides $\sim$8Hz motion data, capturing accelerations and angular velocities that may influence antenna pointing accuracy.

\subsection{Software Architecture}
Synchronizing diverse data streams requires careful architectural decisions. Our ROS-based framework treats time synchronization as a first-class concern. Each sensor node timestamps data using hardware triggers when available, falling back to kernel-level timestamps for software-triggered sensors. A dedicated synchronization node aligns all streams using a combination of GPS time (when available) and local clock correlation, achieving sub-millisecond alignment accuracy across all modalities.

The Starlink data collection presents unique challenges as the terminal doesn't expose a direct API. We employ a multi-layered approach: for satellite tracking and constellation data, we use LEOViz~\cite{ahangarpour2024trajectory, LEOViz2025}, which handles the parsing of Starlink's gRPC status interface and provides 1Hz updates on satellite positions, azimuth/elevation angles, signal quality, and connection status. LEOViz visualizes this data in real-time and we record its output during our experiments.

Data storage must handle high-rate sensor streams while maintaining queryability. We adopt a hybrid approach: high-frequency sensor data streams directly to ROS bag files, preserving full temporal resolution. Communication metrics and derived features are written to time-series databases for efficient analysis. A post-processing pipeline aligns all data sources, producing unified HDF5 files that researchers can analyze without wrestling with format conversions.

%% file: Sections/4_Dataset.tex
\section{Dataset Description}
Our dataset provides synchronized multi-modal sensor data from the Starlink Robot platform, capturing the complete context of mobile satellite communication. The dataset is organized as ROS bag files with extracted CSV files for communication metrics, enabling researchers to analyze the relationship between physical conditions and communication performance.

\begin{figure}[t]
    \centering
    \includegraphics[width=0.48\textwidth]{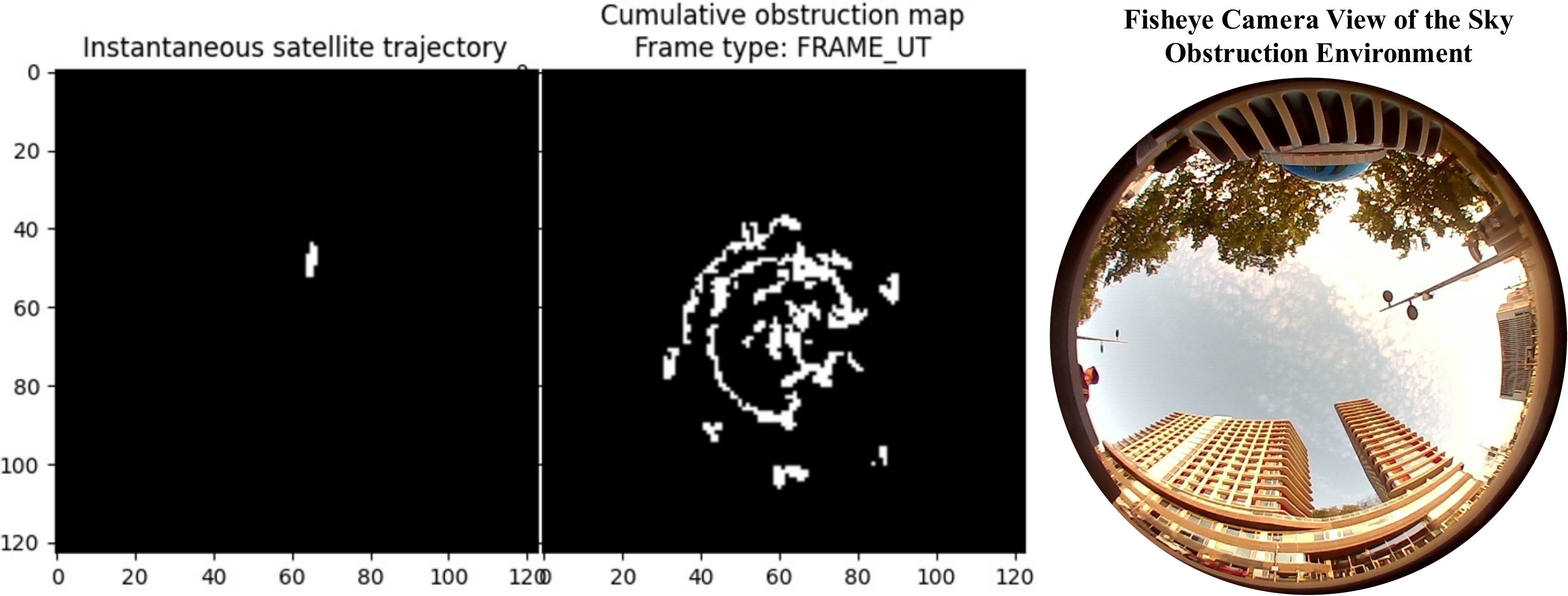}
    \caption{Starlink terminal's obstruction detection output (left) and Dual-view obstruction analysis showing fisheye camera sky visibility (right), demonstrating real-time obstruction mapping capabilities. These are visualized by LEOViz~\cite{ahangarpour2024trajectory, LEOViz2025}.}
    \label{fig:Obstruction}
\end{figure}
\textbf{Obstruction Detection Data.} As shown in Figure~\ref{fig:Obstruction}, we provide dual obstruction detection data. The fisheye camera captures 1920×1080 resolution images at $\sim$15 Hz with a 185-degree field of view, enabling sky visibility analysis. Additionally, we record the Starlink terminal's internal obstruction detection output, which provides the system's own assessment of signal blockage in a black visualization format. Both data streams are time-synchronized, allowing comparison between visual obstruction and the terminal's detection algorithms.

\begin{figure}[t]
    \centering
    \includegraphics[width=0.48\textwidth]{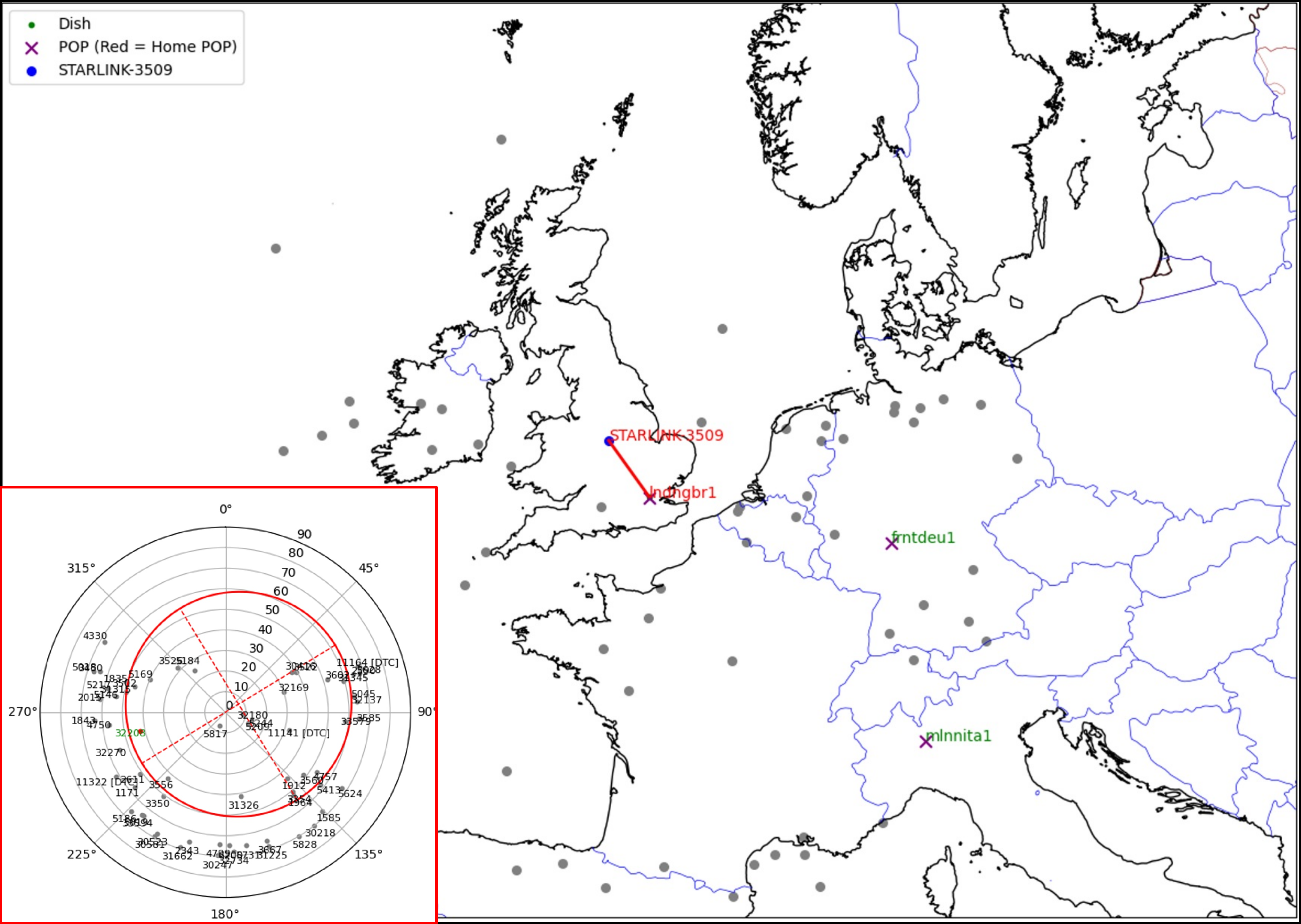}
    \caption{Satellite constellation visualization using LEOViz~\cite{ahangarpour2024trajectory, LEOViz2025}, displaying active Starlink satellites' positions, elevation angles, and connection status relative to the robot's location during data collection.}
    \label{fig:Satellite}
\end{figure}
\textbf{Satellite Tracking Information.} Figure~\ref{fig:Satellite} illustrates the real-time satellite tracking data collected at 1 Hz. For each visible Starlink satellite, we record azimuth, elevation, signal strength, and connection status. This data includes satellite IDs and constellation geometry, enabling analysis of handover patterns and satellite selection behavior during mobile operation.

\begin{figure}[t]
    \centering
    \includegraphics[width=0.48\textwidth]{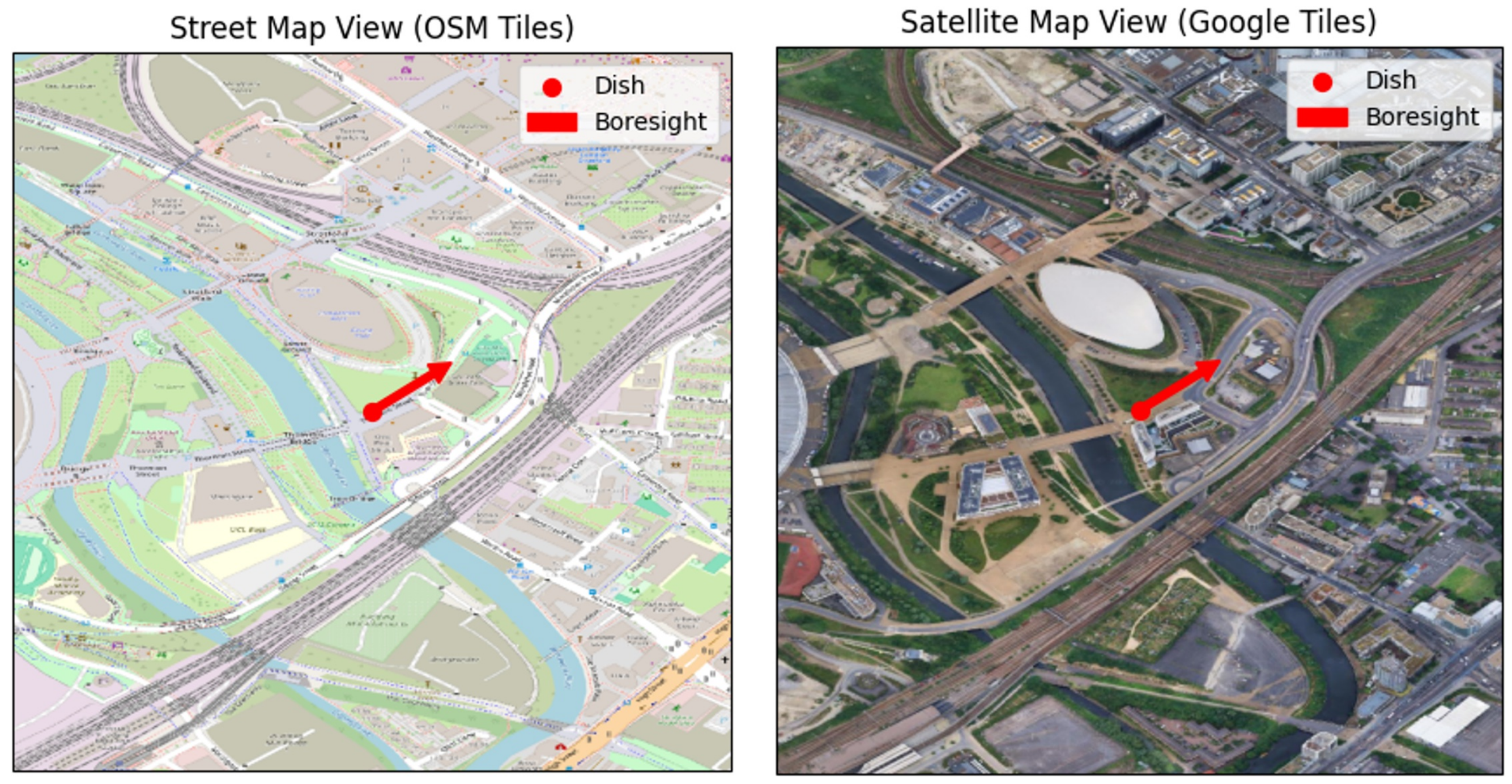}
    \caption{Google map view of the localization utilized LEOViz~\cite{ahangarpour2024trajectory, LEOViz2025}.}
    \label{fig:GoogleMap}
\end{figure}
\textbf{Location and Path Data.} The dataset includes GPS positioning data at 1 Hz, providing global coordinates of the robot's trajectory. As visualized in Figure~\ref{fig:GoogleMap}, our experimental paths cover diverse urban environments including open areas, tree-lined streets, and areas with varying building density. The GPS data is supplemented with wheel odometry at $\sim$8 Hz for improved position accuracy.

\begin{figure}[t]
    \centering
    \includegraphics[width=0.48\textwidth]{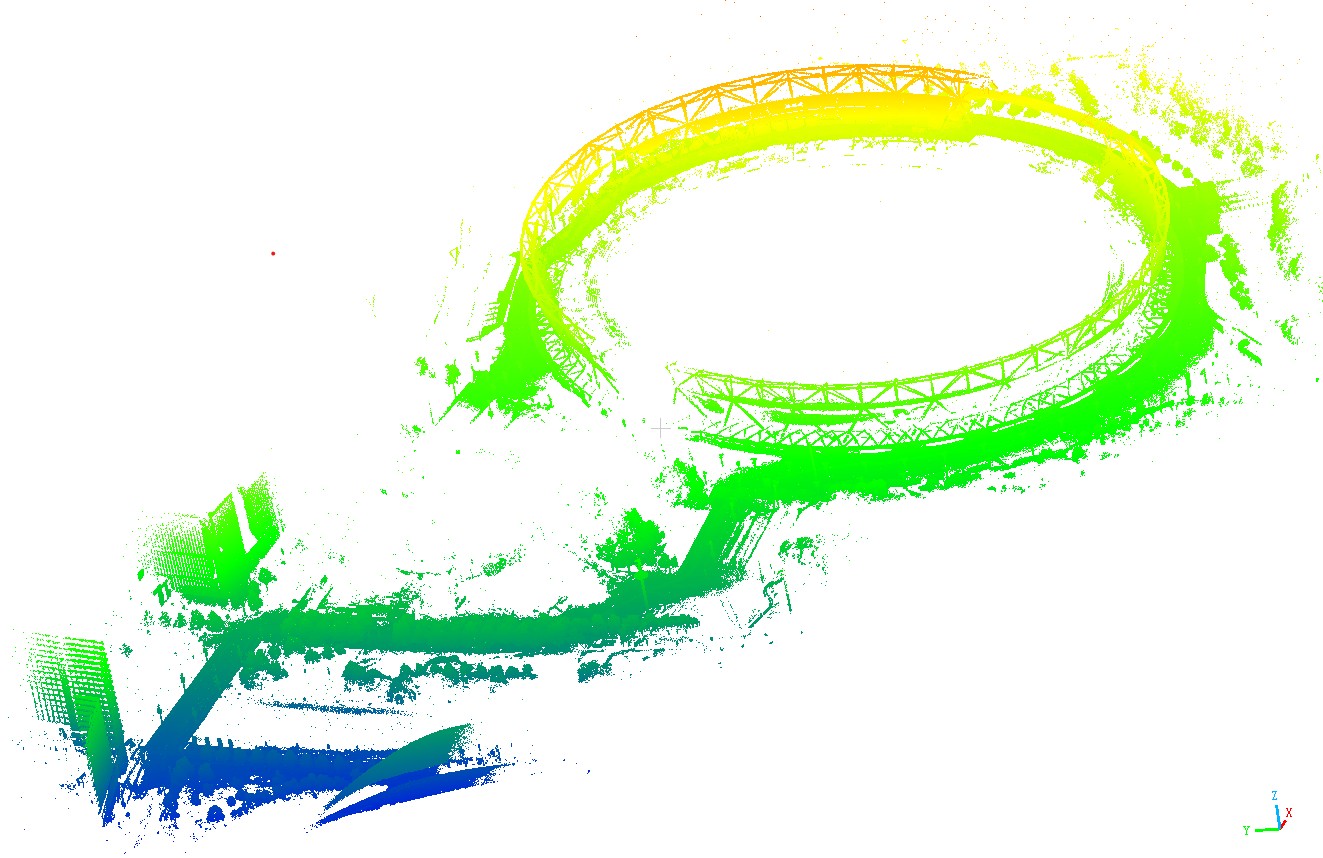}
    \caption{LiDAR-based 3D point cloud visualization capturing environmental geometry around the robot, enabling precise obstruction detection.}
    \label{fig:Map}
\end{figure}
\textbf{3D Environmental Mapping.} The Livox Mid-360 LiDAR captures 360-degree point clouds at $\sim$25 Hz with ranging accuracy of 0.05m up to 40 meters. Figure~\ref{fig:Map} shows example point cloud data revealing environmental geometry including buildings, trees, and other obstacles. The LiDAR data enables 3D reconstruction of the robot's surroundings and correlation with communication performance.

\begin{figure}[t]
    \centering
    \includegraphics[width=0.48\textwidth]{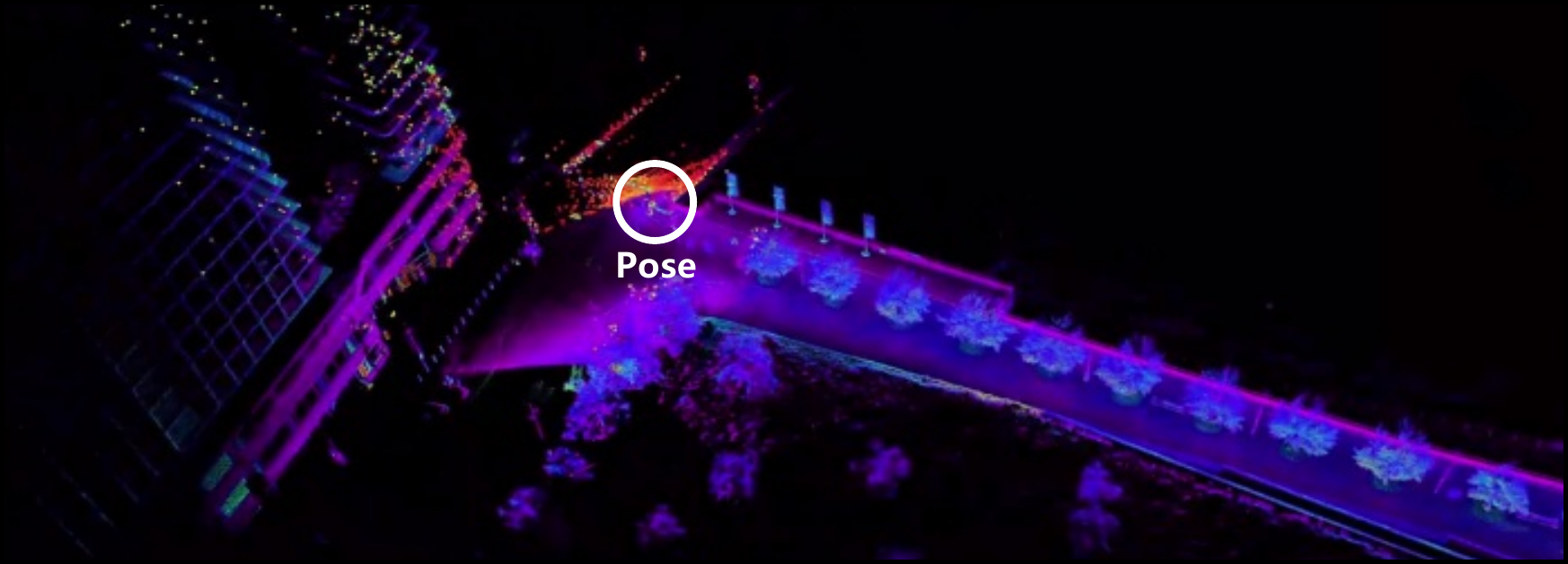}
    \caption{Robot pose and motion data indicating position, velocity, orientation, and trajectory information synchronized with communication performance metrics}
    \label{fig:Pose}
\end{figure}
\textbf{Robot Motion and Pose Data.} Figure~\ref{fig:Pose} presents the comprehensive motion data captured by our platform. The onboard IMU provides $\sim$8 Hz measurements of linear acceleration and angular velocity in three axes. Combined with wheel encoder data, we provide complete 6-DOF pose estimation including position, orientation, velocity, and acceleration. This high-frequency motion data captures vibrations, turns, and speed variations during experiments.

\begin{figure}[t]
    \centering
    \includegraphics[width=0.48\textwidth]{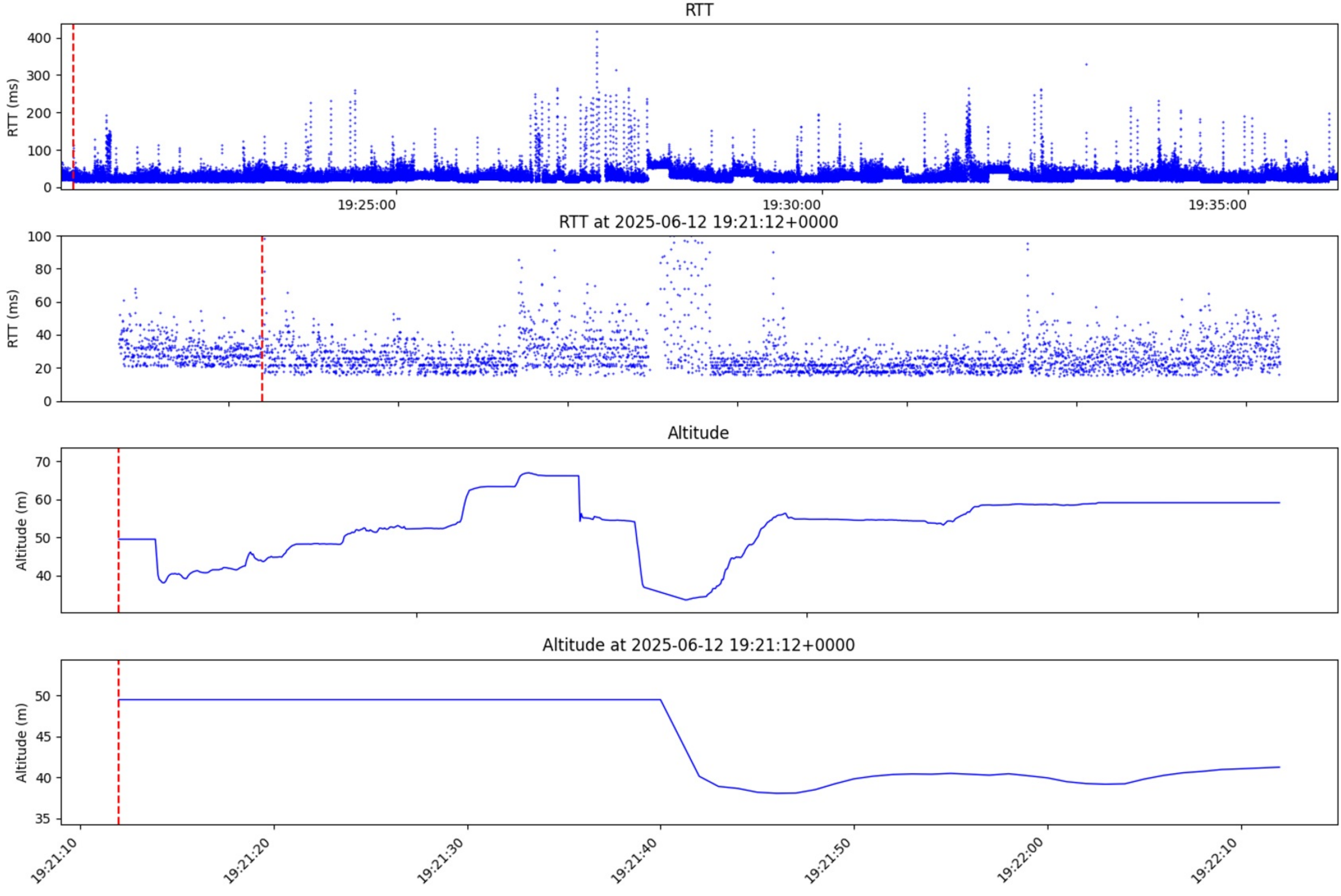}
    \caption{Multi-metric Starlink communication performance indicating RTT variations and altitude profiles during mobile operation.}
    \label{fig:DatasetPlot}
\end{figure}
\textbf{Communication Performance Metrics.} The core of our dataset is the Starlink communication measurements shown in Figure~\ref{fig:DatasetPlot}. We collect with LEOViz: (1) Terminal-reported statistics at 1 Hz including downlink/uplink throughput, RTT, SNR, and obstruction state; (2) Active network measurements with ICMP probes at 10 Hz to multiple servers; (3) TCP/UDP throughput tests every 30 seconds. All communication data is timestamped and synchronized with sensor data.

\textbf{Data Format and Organization.} The dataset is structured as follows - 
 \textit{ROS bags}: Raw sensor data including camera images, LiDAR point clouds, IMU measurements, and GPS coordinates;
 \textit{CSV files}: Extracted communication metrics, satellite tracking data, and robot pose information;
 \textit{Metadata}: Experimental conditions, calibration parameters, and time synchronization information;
 \textit{Timestamps}: All data streams are synchronized using GPS time when available, with local clock correlation for continuous alignment.

Each experimental session includes data from different environmental conditions (open areas, tree-covered paths) and motion patterns (stationary, various speeds). The complete dataset with documentation and processing tools is available for download, providing researchers with the raw data necessary to develop and validate mobile satellite communication algorithms.

%% file: Sections/5_Preliminary_Analysis.tex
\section{Preliminary Analysis}
This preliminary analysis focuses on two critical factors: the impact of movement velocity on satellite connectivity, and variations in communication performance as mobile devices traverse diverse environments. These findings highlight unique challenges inherent to mobile satellite communication that distinguish it from traditional fixed deployments.

\subsection{The Velocity Impact on Communication Performance}
We analyzed communication performance at two different velocities: slow movement (approximately 0.8 m/s) and fast movement (approximately 2.0 m/s). Our findings reveal that motion speed has minimal impact on Starlink communication performance within typical pedestrian velocities.
\begin{figure}[t]
    \centering
    \includegraphics[width=0.48\textwidth]{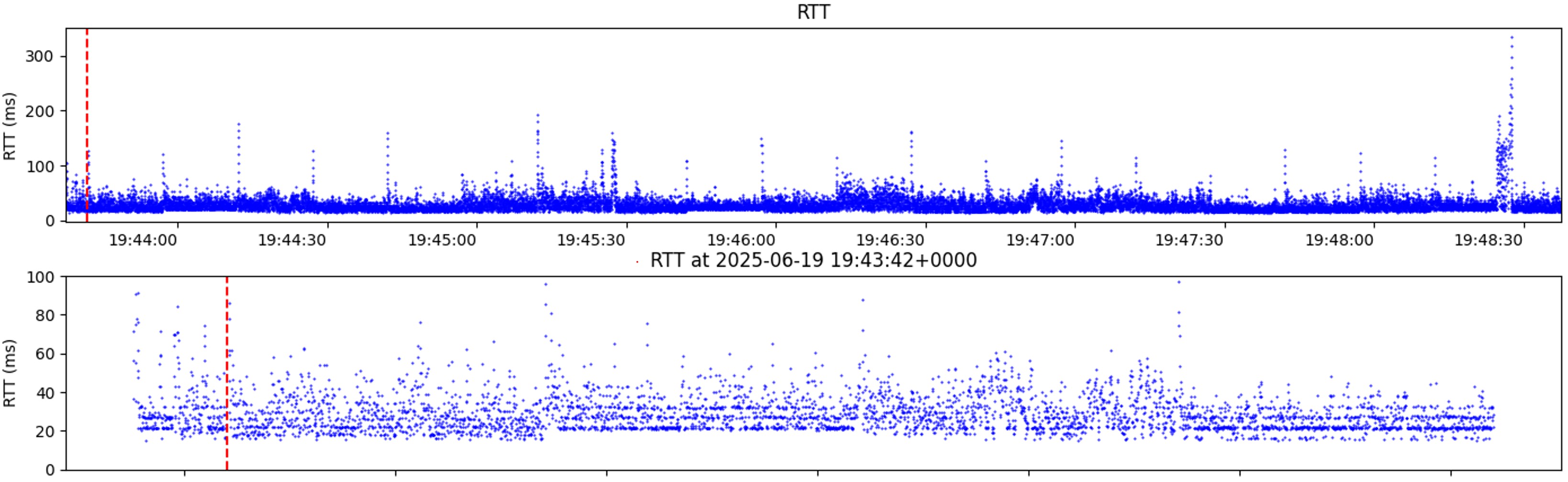}
    \caption{RTT performance during low-speed movement showing communication stability and handover patterns while maintaining slow velocity.}
    \label{fig:low}
\end{figure}
\begin{figure}[t]
    \centering
    \includegraphics[width=0.48\textwidth]{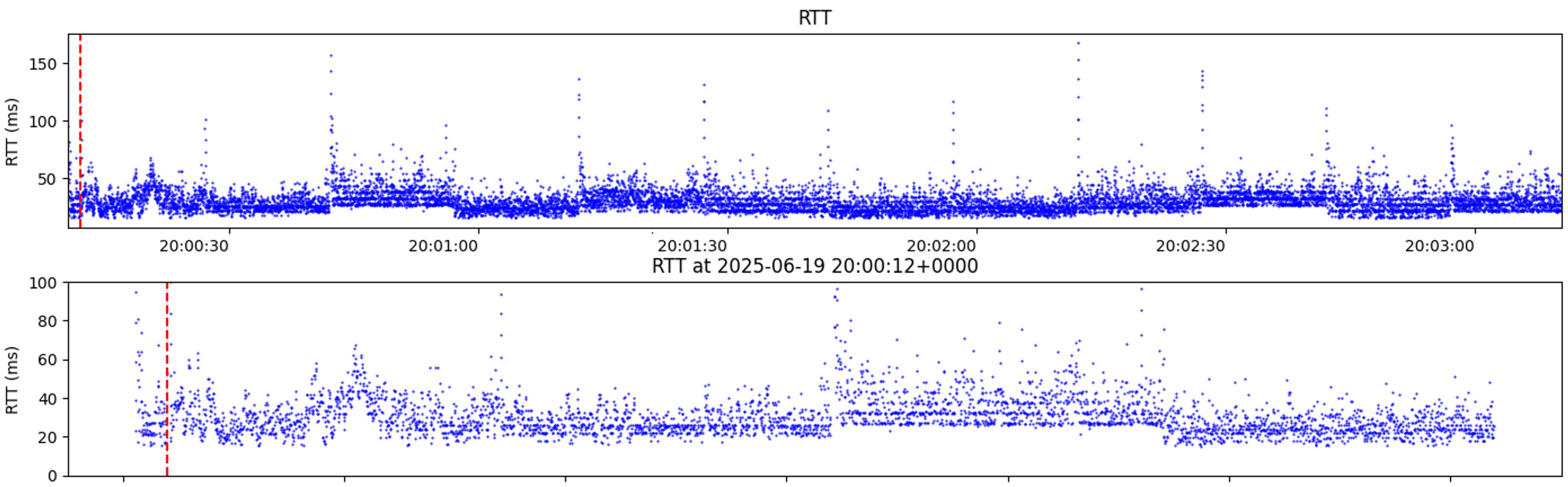}
    \caption{RTT performance during fast-speed movement (2.0 m/s) demonstrating minimal velocity impact on communication quality with slight increase in latency variance.}
    \label{fig:fast}
\end{figure}

As shown in Figures~\ref{fig:low} and~\ref{fig:fast}, RTT measurements remain remarkably stable across both speed conditions, with latency concentrated in the 35-45 ms range. Satellite handovers maintain their predictable 15-second intervals regardless of velocity, manifesting as clear step changes in RTT. While faster movement introduces slightly increased RTT variance—with occasional spikes reaching 60-70 ms—these variations remain within acceptable bounds for most applications.

The minimal performance difference between velocity conditions suggests that Starlink's phased array antenna effectively compensates for motion at pedestrian speeds through electronic beam steering. This finding indicates that for applications involving pedestrian-speed mobility (delivery robots, portable terminals, walking users), velocity itself is not a primary concern for Starlink performance. Instead, as we demonstrate next, environmental factors dominate the mobile user experience.

\subsection{The Environmental Impact on Mobile Satellite Communication}
Mobile devices must continuously adapt to rapidly changing environmental conditions encountered in real-world scenarios. Our data captures the performance of a mobile robotic platform across two representative urban environments, demonstrating the profound influence of environmental dynamics on satellite communication.

\begin{figure}[t]
    \centering
    \includegraphics[width=0.48\textwidth]{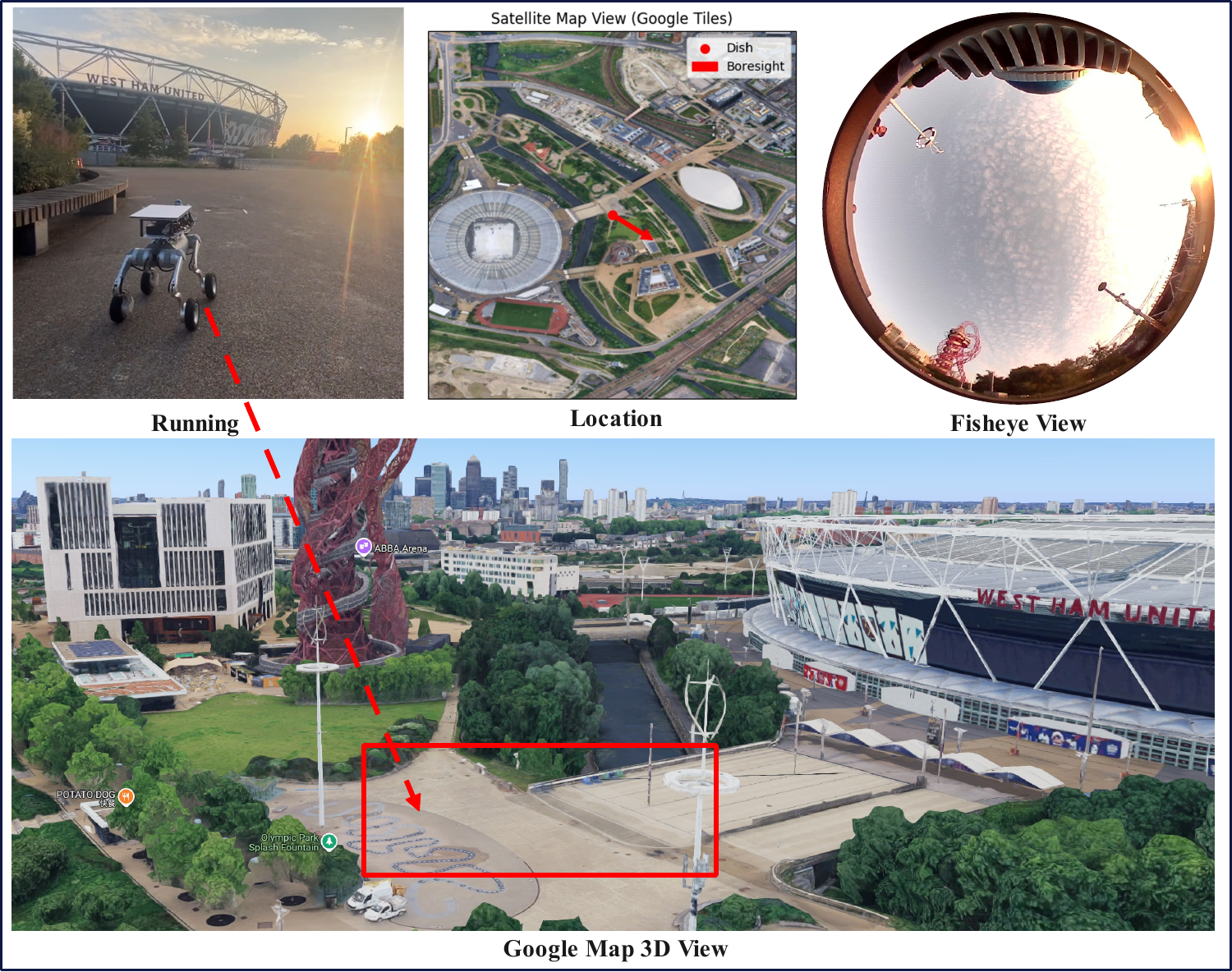}
    \caption{Environmental context and the robot running scenario in open area showing robot deployment location, surrounding infrastructure, and clear sky conditions from multiple viewpoints.}
    \label{fig:BaselineCommunicationScenario}
\end{figure}

\begin{figure}[t]
    \centering
    \includegraphics[width=0.48\textwidth]{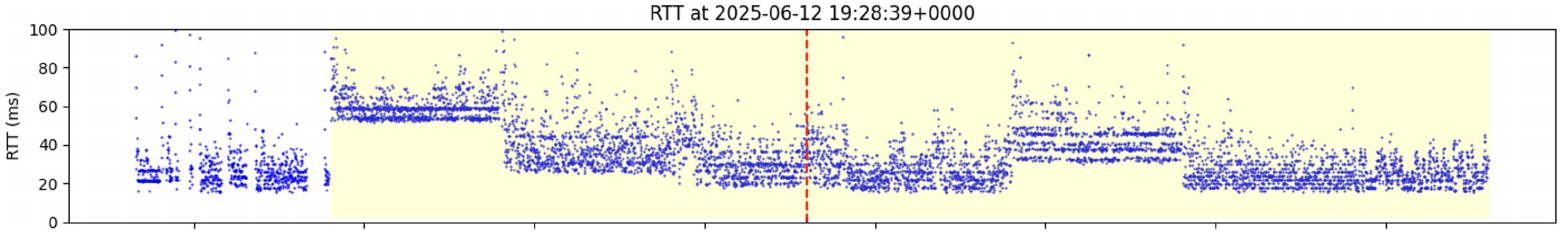}
    \caption{Extended RTT measurements in open environment demonstrating periodic satellite handover patterns and baseline performance characteristics without environmental obstructions.}
    \label{fig:BaselineCommunication}
\end{figure}

In open environments (Figure~\ref{fig:BaselineCommunicationScenario}) - such as the wide roads surrounding a stadium - Starlink communication reflects the core characteristics of LEO satellite systems. As illustrated in Figure~\ref{fig:BaselineCommunication}, RTT (round-trip time) measurements exhibit periodic fluctuations, generally ranging from 20 to 40 ms, corresponding to the satellite handover process. Starlink satellites typically perform handovers approximately every 15 seconds due to their rapid orbital motion. While these handovers are temporally near-seamless, they produce distinct step changes in RTT. In open settings, this pattern is relatively consistent, though latency characteristics vary across different satellites.

\begin{figure}[t]
    \centering
    \includegraphics[width=0.48\textwidth]{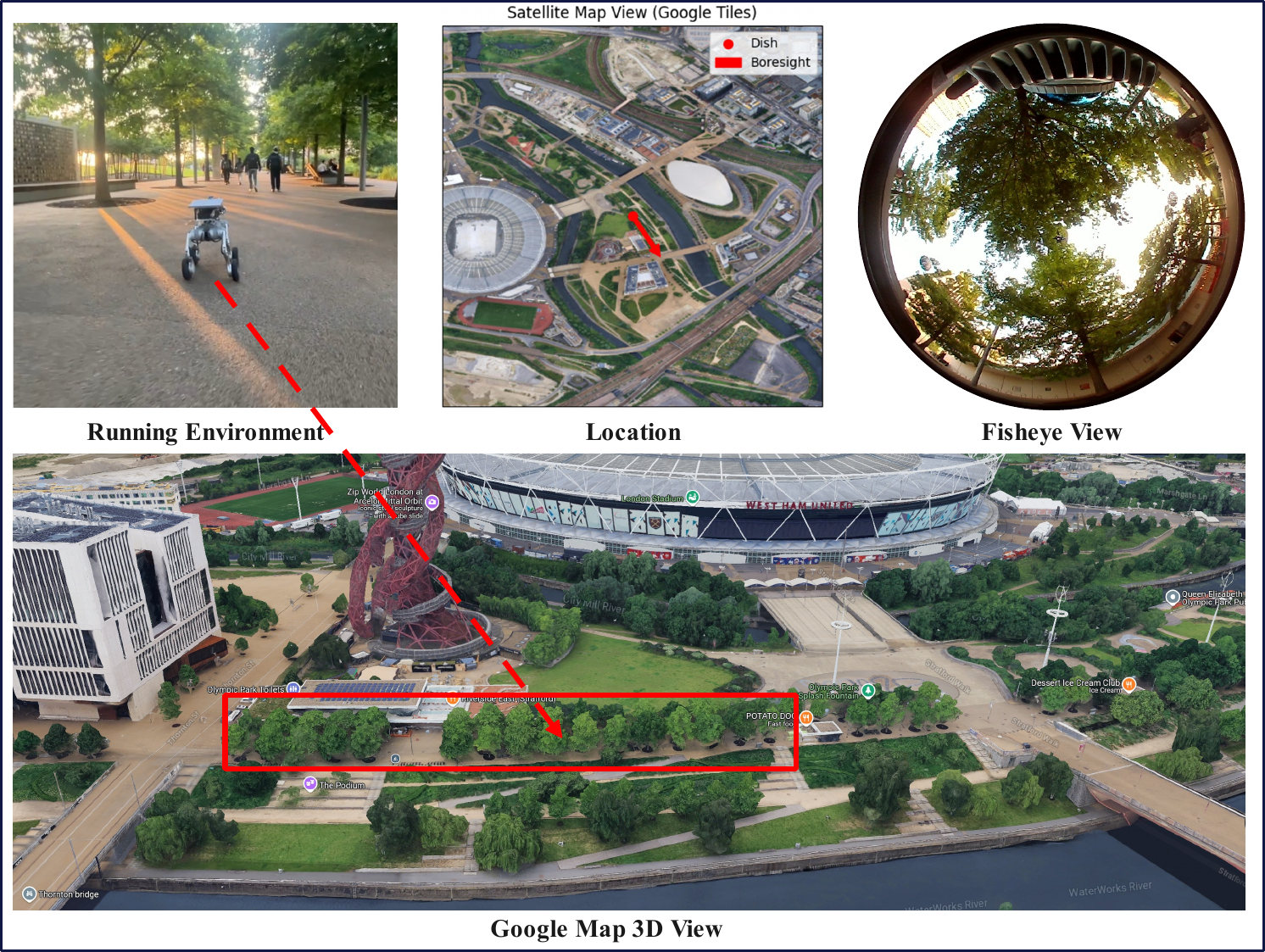}
    \caption{Environmental context and the robot running scenario in tree-covered area showing robot navigation through foliage-dense paths with limited sky visibility.}
    \label{fig:TreeCommunicationScenario}
\end{figure}
\begin{figure}[t]
    \centering
    \includegraphics[width=0.48\textwidth]{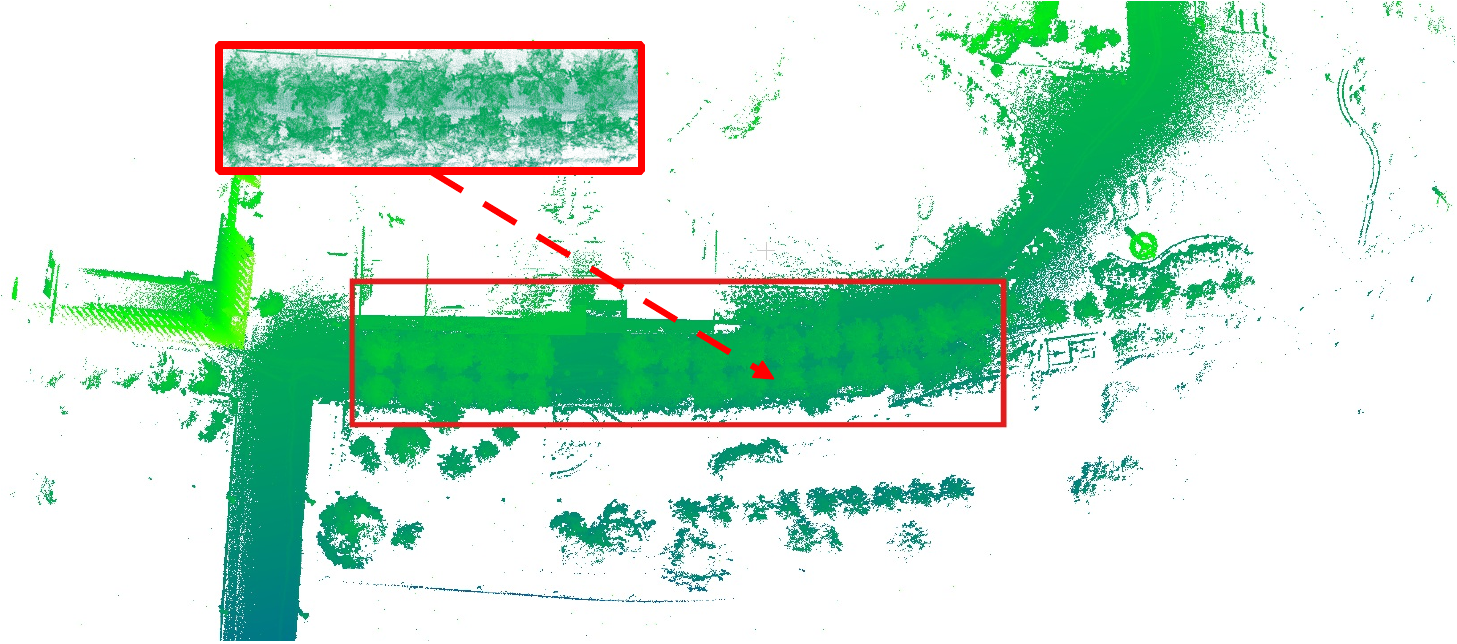}
    \caption{LiDAR point cloud visualization of tree-covered environment illustrating canopy density and potential signal obstruction patterns affecting satellite communication.}
    \label{fig:TreeCommunicationPoint}
\end{figure}
\begin{figure}[t]
    \centering
    \includegraphics[width=0.48\textwidth]{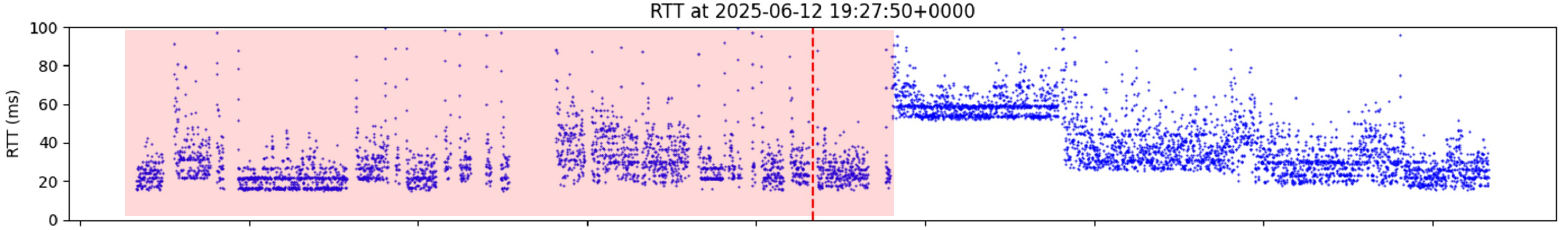}
    \caption{Communication performance in tree-covered environment demonstrating increased RTT instability and frequent spikes due to limited satellite visibility.}
    \label{fig:TreeCommunication}
\end{figure}

In contrast, forested environments (Figure~\ref{fig:TreeCommunicationScenario}) introduce significantly greater variability. Tree-lined streets limit sky visibility and create a dynamic communication environment. As the robot moves along these paths, intermittent canopy gaps continuously alter the terminal's satellite viewing angles. As illustrated in Figure~\ref{fig:TreeCommunication}, RTT data reveals substantial instability, with frequent spikes reaching 40–100 ms. This degradation is not simply the result of signal attenuation; rather, it reflects the compounded effects of handover mechanisms in scenarios constrained by visibility. When fewer satellites are in view, the terminal's handover options diminish, potentially leading to connections with suboptimal satellites or brief service interruptions.

Of particular interest are the sharp changes in communication characteristics during environmental transitions. When the device moves from forested areas into open zones or into building shadows, the system maintains connectivity and transitions to newly available satellites. However, communication performance changes markedly. Our data show significant shifts in RTT baseline, fluctuation amplitude, and overall stability before and after such transitions. These dynamic variations—absent in fixed deployments—are routine for mobile platforms.

%% file: Sections/6_Final.tex
\section{Research Vision and Community Impact}
The Starlink Robot platform transforms how researchers approach mobile satellite communication challenges. By providing synchronized multi-modal data, our platform enables entirely new research directions previously impossible with stationary measurements or uncontrolled mobile experiments. Researchers can now develop motion-aware protocols that anticipate and adapt to environmental changes, design predictive models that maintain quality of service during transitions, and create intelligent path planning algorithms that consider communication constraints alongside traditional navigation objectives.

Our dataset's comprehensive nature empowers the community to tackle fundamental questions in mobile satellite networking. The correlation between sky visibility, motion patterns, and performance metrics allows researchers to move beyond empirical observations to develop theoretical models of satellite-based mobile communications. This understanding is crucial as the industry shifts from best-effort connectivity to guaranteed service levels for critical applications like autonomous vehicles and emergency services. The platform's reproducibility ensures that innovations can be validated across diverse environments and conditions, accelerating the path from research to deployment.

Furthermore, the modular architecture invites community extensions that will shape the future of ubiquitous connectivity. As new satellite constellations launch and devices gain multi-network capabilities, researchers can leverage our framework to study heterogeneous network integration, develop seamless handover mechanisms, and optimize energy consumption across terrestrial and satellite links. By establishing this common measurement infrastructure, we enable the community to build upon each other's work systematically, fostering collaborative progress toward truly global, always-on connectivity.
\section{Conclusion}
This paper introduces the Starlink Robot, the first platform designed specifically for studying satellite communication under controlled mobility. Our multi-modal dataset reveals that motion and environmental dynamics significantly impact satellite performance in ways that static measurements cannot capture. By open-sourcing both the platform design and collected data, we provide the research community with essential tools for developing the next generation of mobile satellite systems.

\section{Acknowledgements}
We thank the developers of LEOViz~\cite{ahangarpour2024trajectory, LEOViz2025} for providing the open-source satellite tracking and visualization tool that enabled the satellite data collection component of our research. LEOViz was instrumental in capturing real-time Starlink satellite positions, elevation angles, and connection status during our mobile experiments. We also acknowledge the open-source robotics community for the ROS framework and associated tools that facilitated our multi-modal data synchronization. 

This work was supported by the UKRI Future Leaders Fellowship [MR/V025333/1] (RoboHike).